\documentclass[runningheads]{llncs}

\usepackage{eccv}

\usepackage{eccvabbrv}

\usepackage{graphicx}
\usepackage{wrapfig}
\usepackage{booktabs}
\usepackage{caption} 
\usepackage[marginal]{footmisc}
\usepackage{colortbl}
\definecolor{Gray}{gray}{0.9}
\usepackage{array,multirow,graphicx}
\usepackage[accsupp]{axessibility}  
\usepackage{marvosym}

\usepackage{hyperref}

\usepackage{orcidlink}

\begin{document}

\title{VUOD: A Versatile Unsupervised Outlier Detection Framework for Natural, Industrial, Medical Images and Beyond}

\titlerunning{VUOD}

\author{Zhonghang Liu\inst{1}\orcidlink{0009-0007-6589-7196} \and
Siyuan Chen\inst{1}\orcidlink{0009-0007-2655-4502} \and
Jingwen Yu\inst{2}\orcidlink{0000-0003-3336-3935} \and
Changshuo Wang\inst{3}\orcidlink{0000-0002-4056-4922} \and
Kunyang Li\inst{4}$^{\textrm{$\ddagger$}}$\orcidlink{0009-0009-1808-8219} \and
Jiangbo Lu\inst{1}$^{\textrm{\Letter}}$\orcidlink{0000-0002-0048-3140}
}

\authorrunning{Z.~Liu et al.}

\institute{SmartMore Corporation, Shenzhen, China \and
The Hong Kong University of Science and Technology, Hong Kong SAR, China  \and
University College London, London, UK \and Rice University, Houston, USA \\
\email{jiangbo.lu@gmail.com}
} 

\maketitle
\renewcommand{\thefootnote}{\fnsymbol{footnote}}
\footnotetext[0]{\textsuperscript{$\ddagger$} Work done during an internship at SmartMore,
\textsuperscript{\Letter} Corresponding author.}
\renewcommand{\thefootnote}{\arabic{footnote}}

\begin{abstract}
  Unsupervised outlier detection that automatically identifies whether visual systems involve anomalous images is a significant research topic. However, most approaches are limited to natural images because they rely on frozen and pre-trained feature extractors. So that their performance cannot be maintained in practical scenarios, especially for industrial inspection and medical imaging. In this work, we first revisit this task and then introduce a versatile unsupervised outlier detection framework to enrich the application domains. The core idea of this framework is to improve feature discriminativess via exploiting intrinsic distribution priors. Evaluated on 3 domains and 14 benchmark datasets, our proposed solution achieves state-of-the-art performance and significantly outperforms existing methods. More importantly, we show its plug-and-play property that can be integrated into diverse visual applications to improve their robustness, such as image classification and 3D reconstruction. Our code is available at \href{https://github.com/zhliu-uod/VUOD}{https://github.com/zhliu-uod/VUOD}.
  
\end{abstract}

\section{Introduction}
\label{sec:intro}
\begin{figure}[t]
\centering
\includegraphics[width=0.98\textwidth]{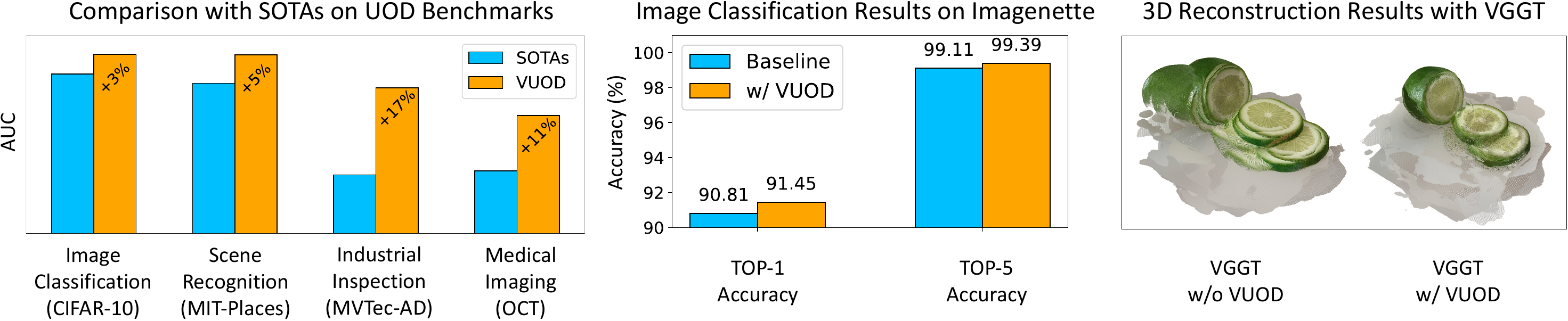} 
\caption{\textbf{The highlights of our proposed VUOD framework.} Left plot: VUOD significantly outperforms existing SOTAs on UOD benchmark datasets and real-world applications. More importantly,
VUOD can be seamlessly integrated into various downstream visual tasks, such as image classification (middle plot) and 3D reconstruction with VGGT \cite{wang2025vggt} (right plot), to improve their performance and robustness.}
\label{fig 1} 
\end{figure}

In the realm of image data, outliers are inevitable. For example, the out-of-distribution data points (e.g., trucks within the ``car'' category) and defective manufacturing products (e.g., ``screw'' with a scratch). Unsupervised outlier detection (UOD), aiming to automatically identify anomalous images, is a significant research topic. However, since recent methods \cite{lin2021shell, liu2024rethinking, liu2025flexuod} rely on natural image pretrained feature extractors \cite{he2016deep, radford2021learning}, their performance will be dramatically affected in various practical domains. In this work, we introduce a versatile unsupervised outlier detection (VUOD) framework, which enables UOD methods to be effective on scene recognition \cite{zhou2017places}, industrial inspection \cite{bergmann2019mvtec, bergmann2022beyond} and medical imaging \cite{fernando2021deep}, also improving the robustness of image classification \cite{imagenet} and 3D reconstruction \cite{wang2025vggt}, as illustrated in Fig. \ref{fig 1}.

To this end, the core idea is to improve the feature discriminativeness among different domains. From this perspective, we will mainly focus on exploiting reliable pseudo labels, i.e., self-supervision that transforms the current unsupervised task into a supervised problem, and leverage contrastive learning to enhance the separability of image feature representation.
Previously, this was usually generated by threshold learners \cite{lin2021shell, liu2024rethinking} that learn a decision boundary on a series of outlier scores given by an external unsupervised outlier detector. However, this paradigm suffers from a significant \text{inconsistency} problem caused by the independence between threshold learners and outlier scoring methods.

In this study, we present a novel perspective by leveraging the prior knowledge of the target dataset without bringing in any external supervision signals. Two intrinsic priors, local separability and global separability, derived from our theoretical and empirical observations, are introduced. Specifically, in the local region, we utilize a fine-grained clustering approach to split the target set into a series of clusters, while in the global space, we explore a global separability prior that performs as a binary classifier to annotate each local cluster.

To evaluate the performance of our proposed solution, we subject it to rigorous and broad testing, which involves 3 domains (natural, industrial and medical) and 14 benchmark datasets.
Our experimental results demonstrate that it achieves the overall state-of-the-art (SOTA) results and is also promising in terms of efficiency.
The main contributions of this work can be summarized as: 
\begin{itemize} 
\item First, we comprehensively revisit and analyze the UOD task, including its definition and real-world applications.
This could be beneficial for further fair evaluation and inspires future research perspectives.
\item We propose a versatile UOD framework that offers several advantages: \\
$\operatorname{(\romannumeral1)}$ Evaluated on standard UOD benchmarks, it achieves SOTA performance and significantly outperforms competitors on challenging datasets. \\
$\operatorname{(\romannumeral2)}$ Evaluated on practical scenarios, it significantly outperforms other UOD baselines, e.g., $16.7\%$ improvement on MVTec-AD, and achieves competitive performance compared with specific defect detectors. \\ 
$\operatorname{(\romannumeral3)}$ It helps improve the robustness of visual tasks like image classification and 3D reconstruction via its significant outlier filtering ability.
\end{itemize} 

\section{Related Work}
\label{sec:related}

\subsection{Unsupervised Outlier Detection}
UOD methods can be divided into two major categories: deep learning-based and statistical-based. Deep approaches \cite{lai2020rsrae,zong2018deep,zhou2017anomaly,wang2019effective,li2021deep} 
attempt to compress image samples into a low-dimensional space and reconstruct them. The underlying assumption is that the model will successfully learn the patterns of inliers, resulting in low reconstruction errors.
In contrast, statistical-based methods utilize discrimination \cite{lin2021shell, lin2022locally, liu2024rethinking, liu2025flexuod} or density \cite{kriegel2008angle, liu2008isolation} strategies to model the inlier manifold. However, most of them significantly rely on a natural image pretrained feature extractor, e.g., ResNet-50 \cite{he2016deep}, and CLIP \cite{radford2021learning}, which limits their ability to generalize into diverse applications \cite{bergmann2019mvtec, mishra2021vt, jezek2021deep, bergmann2022beyond, bao2024bmad}. To address this limitation, we will focus on enhancing feature discriminativeness across varied domains.

\subsection{Feature Discriminativeness Learning}
Mainstream end-to-end self-supervised models \cite{wang2019effective, 9816125, huyan2022unsupervised} typically enhance feature discriminativeness through contrastive-based or pretext-based strategies. They often require carefully designed augmentations and domain-specific heuristics that may not align with the subtle nuances of outlier detection.
In contrast, self-supervision generation (pseudo-labeling) in UOD is typically achieved through the thresholding (threshold learning) \cite{qi2021iterative, friendly2013elliptical, barbado2022rule, thanammal2014effective, hashemi2019filtering, martin2006evaluation, afsari2011riemannian, aggarwal2017introduction, iouchtchenko2019deterministic, barbado2022rule, boente2002influence, zhao2020automating, keyzer1997using, breunig2000lof, van2020evaluation, perini2023estimating} process, which determines a decision boundary based on a series of given outlier scores. 
However, because conventional threshold learning is independent of the underlying outlier detector, and the discriminative quality of the resulting outlier scores is not guaranteed, i.e., threshold-based methods often perform sub-optimally in practice. To address these limitations, our approach predicts pseudo labels by leveraging intrinsic distribution priors within the feature space.

\section{Revisiting Unsupervised Outlier Detection}
Before proposing our methodology, we discuss the foundational concepts for the UOD task. We first formally define the problem, including its notations and the core objective. Following this, we explore how UOD can be applied across diverse real-world domains and integrated into broader visual perception applications.

\subsection{Problem Definition}
The ``unsupervised'' term in the outlier detection field mainly refers to no labeled data being involved in the training phase. In other words, UOD methods are trained and tested on the same \textit{target dataset} \cite{liu2025flexuod}, which involves $n$ unlabeled images $\mathbf{I} = \mathbf{I}_{in} \cup \mathbf{I}_{out}$, where $\mathbf{I}_{in}$, $\mathbf{I}_{out}$ refer to inliers and outliers, respectively. 
The outlier ratio (contamination factor) $\gamma  \in (0, 1)$ is $\frac{\#  \mathbf{I}_{{out}}}{\#  \mathbf{I}_{{in}}+\#  \mathbf{I}_{{out}}}$, which is also unknown. 
By default, we utilize some pretrained feature extractors, e.g., ResNet-50 \cite{he2016deep} and CLIP \cite{radford2021learning}, to obtain the target features $\mathbf{X} = \mathbf{X}_{in} \cup \mathbf{X}_{out} \in \mathbb{R}^{n \times d}$, where $d$ refers to the dimension. 
Although UOD typically is a binary classification that includes both scoring (predicting outlier scores for each sample) and thresholding (deciding the boundary to predict labels), the thresholding phase will be an ill-posed problem without a separable outlier score distribution \cite{liu2024rethinking}. The core objective of UOD approaches is to learn an outlier score function ${f}(\cdot)$ to quantifies the anomalous probability of each $x_{i \in [1, n]} \in \mathbf{X}$, and the ideal ${f}(\cdot)$ should achieve a $100\%$ ranking accuracy, i.e., $\operatorname{max}(\{f(x_i)| x_i \in \mathbf{X}_{in} \}) < \operatorname{min}(\{f(x_i)| x_i \in \mathbf{X}_{out} \})$. 

\begin{figure*}[t]
\centering
\includegraphics[width=0.98\textwidth]{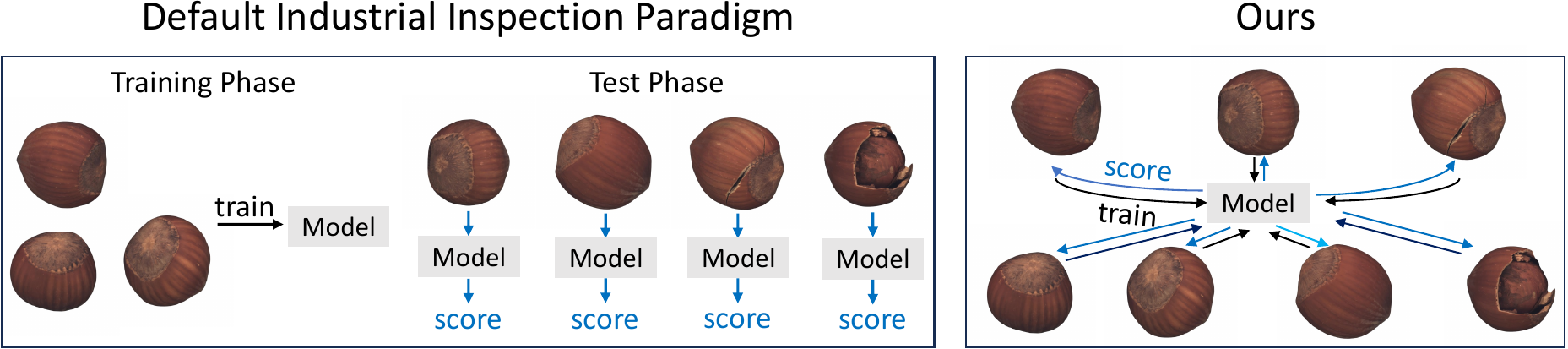} 
\caption{\textbf{Comparison between our paradigm and default industrial inspection.} Left plot: The default unsupervised paradigm for industrial inspection, requiring a training phase on defect-free samples before sequential inference.
Right plot: Our introduced paradigm eliminates the split between training and testing, enabling flexible, simultaneous detection of multiple objects within a single scene.
}
\label{fig 2} 
\end{figure*} 

\subsection{Real-world Applications}
To date, as previously discussed, most UOD methods have focused on detecting anomalous samples in natural image datasets, such as trucks that appear in the ``car'' class. However, UOD has significant value beyond this setting. It plays an important role in various real-world domains, including industrial inspection \cite{bergmann2019mvtec, mishra2021vt, jezek2021deep, zou2022spot, roth2022towards, bergmann2022beyond, liu2024deep} and medical imaging \cite{cai2025medianomaly}. 
\text{Remarkably}, the paradigm of applying UOD methods into industrial inspection (defect detection) is different from default approaches, as shown in Fig. \ref{fig 2}, and our solution demonstrates more flexibility since its unsupervised property.
Moreover, UOD can be integrated as a complementary component in broader visual perception pipelines. For example, in image classification tasks, UOD can be used to remove out-of-distribution samples from the training set, thereby improving model robustness. Similarly, for feed-forward 3D reconstruction pipelines \cite{wang2025vggt} that rely on consistent scene or object imagery, UOD can help filter mismatched or corrupted inputs, leading to more accurate reconstruction results. This will be evaluated in Sec. \ref{sec. 5.4}.

\section{Method}
\label{sec:method}

\begin{figure*}[t]
\centering
\includegraphics[width=0.98\textwidth]{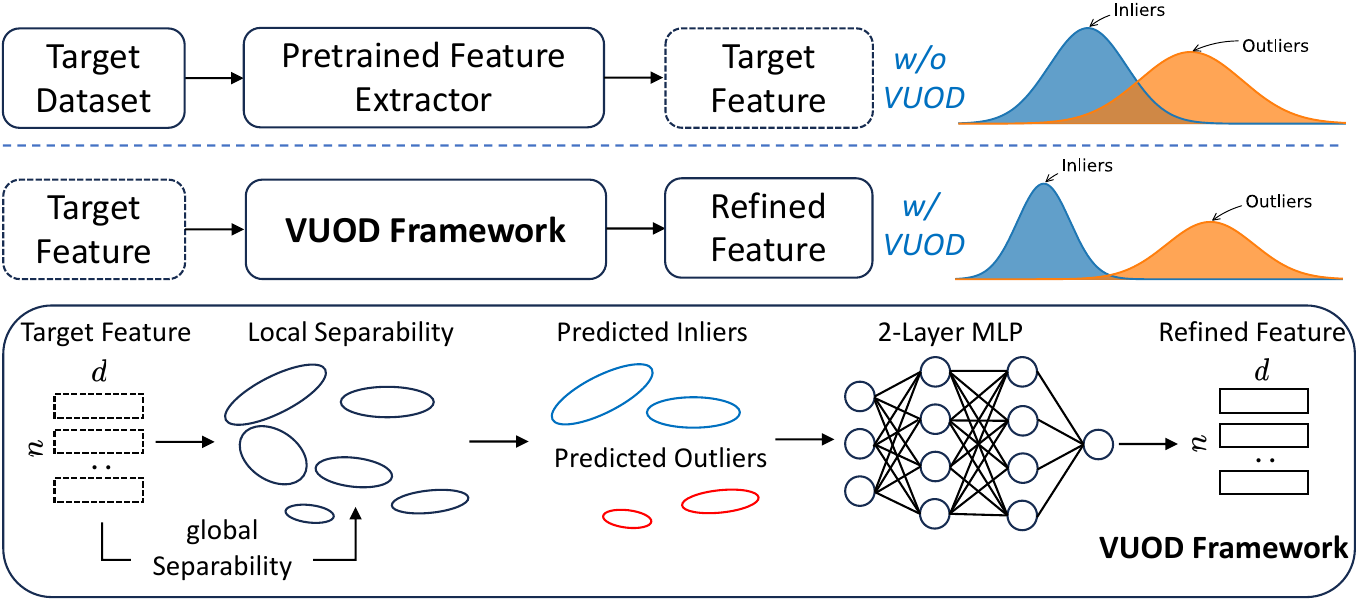} 
\caption{\textbf{The core idea of our proposed VUOD framework.} An unlabeled target dataset is first processed by natural image pretrained feature extractors. We then investigate the intrinsic priors of the feature space (local and global separability), and generate pseudo labels. Based on predicted inliers and outliers, we apply contrastive learning to enhance the discriminativeness of the feature representations.}
\label{fig 3} 
\end{figure*} 

\begin{figure*}[htbp]
\centering
\includegraphics[width=0.98\textwidth]{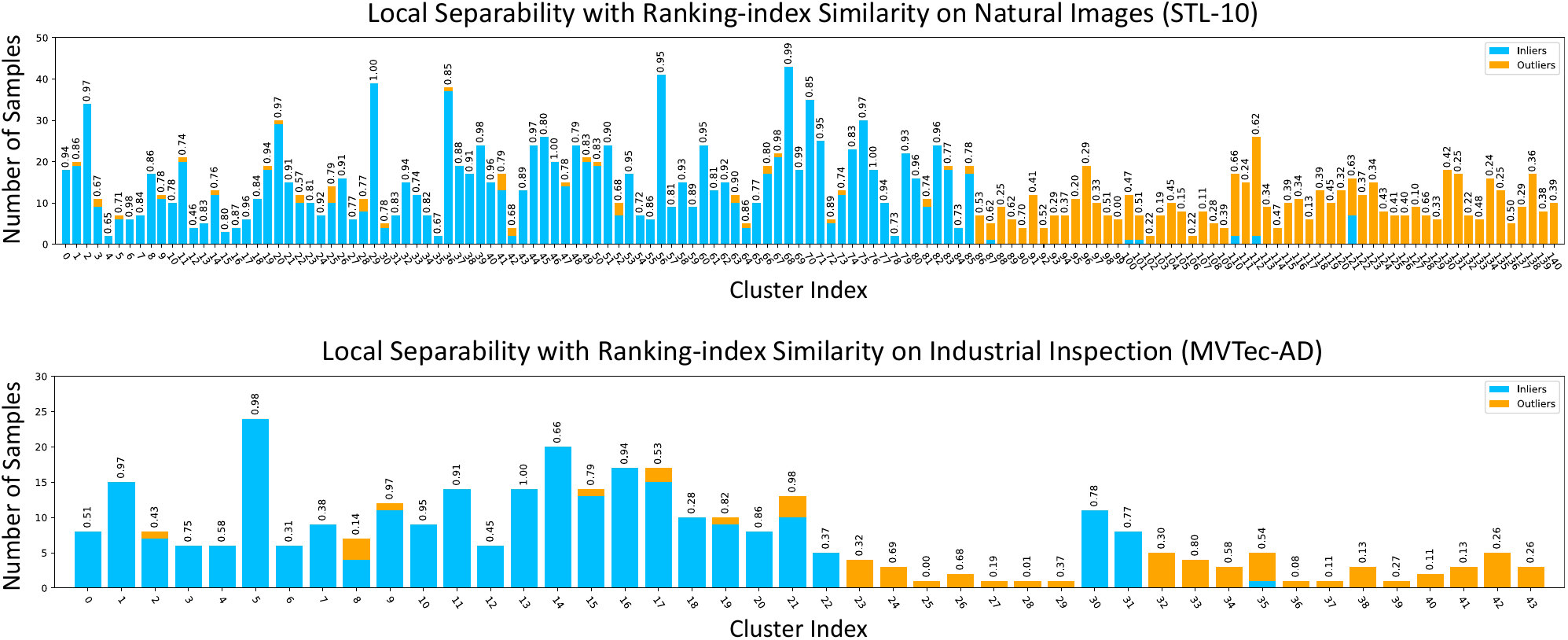} 
\caption{\textbf{The visualization of local separability with ranking-index similarity (Eq. \ref{equ 5})}.
Intuitively, the fine-grained clustering algorithm Affinity Propagation \cite{frey2007clustering} is capable of effectively separating outliers, even when the pretrained feature embeddings exhibit limited separability. In practice, uncontaminated inlier clusters typically show high ranking-index similarity (often larger than $0.9$), while uncontaminated outlier clusters tend to exhibit substantially lower similarity values (often less than $0.5$).}
\label{fig 4} 
\end{figure*}

\subsection{Overview}
To apply UOD approaches into diverse domains, the core idea is to enhance the discriminativeness of feature representation via \textit{pseudo labeling} (self-supervision generation), i.e., splitting both inlier features $\mathbf{X}_{in}^{\prime}$ and outlier features $\mathbf{X}_{out}^{\prime}$ from $\mathbf{X}$, and thereby leveraging \textit{contrastive learning} strategy to enlarge the similarity within inliers and the difference between inliers and outliers in the enhanced feature space. The overview of our framework is illustrated in Fig. \ref{fig 3}.   

\subsection{Separability Priors}
\label{sec 3.2}
We decompose \text{pseudo labeling} into \textcolor{blue}{splitting} and \textcolor{red}{classifying}. Specifically, in the local region, we utilize a fine-grained clustering approach to split the target set into a series of clusters, while in the global space, we explore a separability prior that performs as a supervision signal to classify each local cluster.

\noindent \textbf{Local Separability.}
In the first \textcolor{blue}{splitting} phase, we adopt $\operatorname{Affinity \ Propagation}$ ($\operatorname{AP}$)  
\cite{frey2007clustering}, which will split the target dataset's features $\mathbf{X}$ into a series of fine-grained clusters $\{\mathbf{C}_t \}_{t=1}^{T}$ without pre-defining the clustering number $T$,
where $\mathbf{C}_t$ refers to $t$-th cluster. It proceeds by alternatively updating two matrices: $\operatorname{responsibility}$ and $\operatorname{availability}$. The $\operatorname{responsibility}$ is defined as: 
\begin{equation}
r(x_i, x_k) \leftarrow s(x_i, x_k) - \max_{k^{\prime} \ne k} \{a(x_i, x_{k^{\prime}}) + s(x_i, x_{k^{\prime}}) \},
\label{equ 1}
\end{equation} 
which quantifies how well-suited $x_k$ is to serve as the exemplar (centroid) for $x_i$, relative to other candidate exemplars $x_{k^{\prime}}$. Here $s(\cdot) = -\displaystyle ||\cdot ||_2^2$ refers to the similarity (negative squared euclidean distance), the $\operatorname{availability} a(x_i, x_k) \leftarrow \min (0,\, r(x_k, x_k) + \sum_{i' \notin \{i, k\}} \max(0,\, r(x_{i^{\prime}}, x_k)))$ represents how proper it would be for $x_i$ to pick $x_k$ as its exemplar, considering other points' preference for $x_k$ as an exemplar.
Based on some common sense in the UOD field \cite{ye2023uadb, du2022vos, isaacmedina2024towards, du2023dream, tao2023nonparametric}, 
outliers are typically sparse, with large variance between each other and low similarity to inliers (normal samples). For example, if the inliers are good ``capsules'', outliers could be ``capsules'' with cracks, pokes, scratches, etc.  \cite{bergmann2019mvtec}. Such that we could hypothesize that 
\begin{equation}
\mathbb{E}_{x_i \in \mathbf{X}_{out}, x_j \in \mathbf{X}_{in}}[s(x_i,x_j)] \ll \mathbb{E}_{x_j, x_{j^{\prime}} \in \mathbf{X}_{in}}[s(x_j,x_{j^{\prime}})],
\label{equ 2}
\end{equation}
where $\mathbb{E}[\cdot]$ is the exception. In the responsibility $r(x_i, x_k)$ update step of $\operatorname{AP}$, for an outlier $x_i \in \mathbf{X}_{out}$, there might exist no $x_k$ (whether it is an inlier sample or another outlier) that can provide sufficiently high similarity. As a result, it may not be strongly attracted to any inlier cluster; or if there exists weak similarity among outliers, they may form a small cluster; or if completely isolated, it becomes a singleton cluster. 

Consequently, all these clusters of $\operatorname{AP}$ can be divided into three categories:
$\operatorname{(\romannumeral1)}$ inlier clusters that each cluster mostly contains inliers;
$\operatorname{(\romannumeral2)}$ outliers clusters that each cluster mostly contains outliers;
$\operatorname{(\romannumeral3)}$ mixed clusters that contain both inliers and outliers. The visualization can be found in Fig. \ref{fig 4}, and more visualizations can be found in the Appendix.
Next, we will explore a supervision signal to predict the labels of each cluster and split category $\operatorname{(\romannumeral1)}$ and $\operatorname{(\romannumeral2)}$ from $\mathbf{X}$.

\begin{figure*}[t]
\centering
\includegraphics[width=0.98\textwidth]{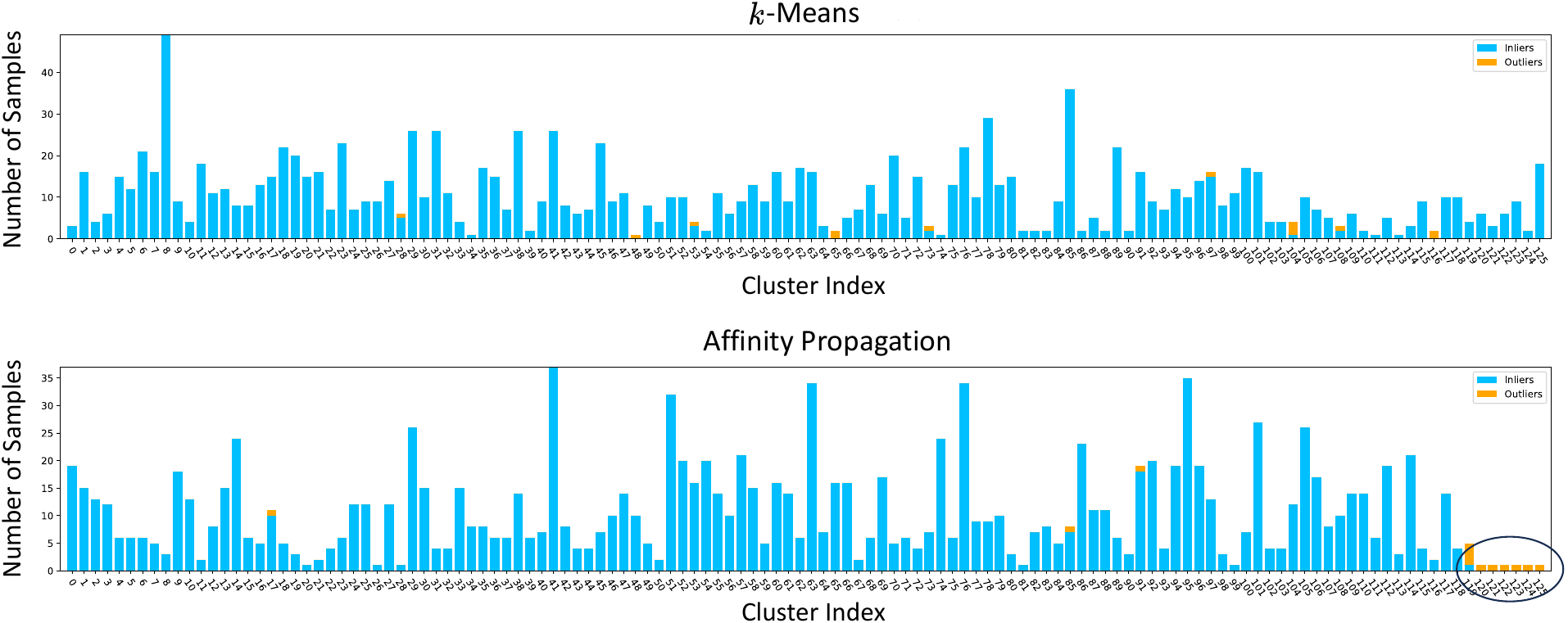} 
\caption{\textbf{The clustering comparison between Affinity Propagation and $k$-Means.} Since $k$-Means requires a predefined number of clusters, which is impractical in unsupervised settings, we adopt the automatically determined number of clusters from Affinity Propagation for a fair comparison. As illustrated, Affinity Propagation consistently performs better than $k$-Means, particularly when outliers constitute only a small fraction of the target dataset.}
\label{fig 5} 
\end{figure*}

\noindent \textbf{Discussion: Why not $k$-Means?}
In this work, we apply $\operatorname{Affinity \ Propagation}$ instead of the widely used
$k$-Means \cite{macqueen1967classification, lloyd1982least, lin2022locally, zhu2024anomaly} to measure local separability. This is because $k$-Means implicitly assumes that the data are generated from several compact and spherical clusters, and it enforces each sample to be assigned to one of them. In other words, there is no mechanism to reject an assignment or leave a sample un-clustered. Consequently, outliers that even deviate significantly from the inlier distribution tend to be absorbed into the nearest cluster, which makes it difficult to separate them. The comparison is visualized in Fig. \ref{fig 5}. Besides, one of the most challenging problems that utilizes $k$-Means is learning $k$ in $k$-Means without any supervision, which is a challenging problem \cite{lin2022locally}.

\begin{figure}[ht]
\centering
\includegraphics[width=0.5\textwidth]{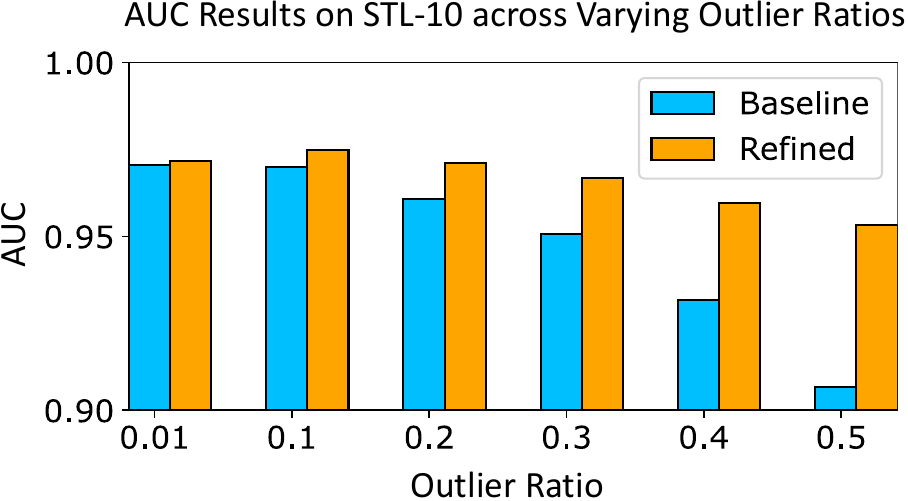}
\caption{\textbf{The AUC results comparison between the refined distance (Eq. \ref{equ 6}) and the baseline (Eq. \ref{equ 4}).}
By removing the outlier impact from the mean computation, the performance of global separability can be maintained across different outlier ratios, i.e., the challenging mean-shift problem is well addressed.
}
\label{fig 6} 
\end{figure}

\noindent \textbf{Global Separability.}
Inspired by some theoretical analysis about the distribution of the target high-dimensional features \cite{lin2018dimensionality, lin2021shell, lin2023distance}, the distance from each sample to the centroid of inliers $m_{\mathbf{X}_{in}}$ could serve as a naturally discriminative outlier score function, i.e., for most $x_i \in \mathbf{X}_{in}$ and $x_j \in \mathbf{X}_{out}$, $\displaystyle ||x_i - m_{\mathbf{X}_{in}} ||2 < \displaystyle ||x_j - m_{\mathbf{X}_{in}} ||2$.
This is because inliers $\mathbf{X}_{in}$ are contained within a $d$-dimensional hypersphere of radius $r$, the volume is denoted as: $\mathbf{V}_d = (\pi^{d / 2} r^d) / \Gamma(1+d / 2)$, where $\Gamma(\cdot)$ refers to the Gamma function. Shrinking $r$ by a small value $\delta$ ($\delta>0$) yields the following limit:
\begin{equation}
\lim_{d\to\infty}  \frac{\mathbf{V}_d((1-\delta) r)}{\mathbf{V}_d} 
=  \lim_{d\to\infty}(1-\delta)^d  \leq \lim_{d\to\infty} e^{-\delta d} = 0.
\label{equ 3}
\end{equation}
As $d \rightarrow \infty$, almost all the volume concentrates on the hypersphere's surface. Given a low outlier ratio $\gamma$ (e.g. $<1\%$), the total feature mean $m_{\mathbf{X}}$ remains highly close to the inlier mean $m_{\mathbf{X}_{in}}$. Therefore,
\begin{equation}
f_{base}(x_i) = \displaystyle || x_i - {m_{\mathbf{X}}} ||^{2}
\label{equ 4}
\end{equation}
provides reliable outlier scores. In contrast, when $\gamma$ is relatively high, the gap between $m_{\mathbf{X}}$ and $m_{\mathbf{X}_{in}}$ becomes larger, which is considered a classic mean-shift problem \cite{reiss2023mean}. To address this issue, we tend to remove some outliers based on an intuitive conclusion: $f(x_i) = \displaystyle || x_i - {m_{\mathbf{X}_{out}}} ||^{2}$ is an \text{invalid} outlier score function. Likewise, if some local $\mathbf{C}_t \in \mathbf{X}_{out}$ are obtained from the above local separability section, $f_t(x_i) = \displaystyle || x_i - {m_{\mathbf{C}_{t}}} ||^{2}$ is also invalid.
This contrastive property inspires us to the ranking-index results between $\{f_{t}(x_i)\}_{i=1}^{n}$ and $\{f_{base}(x_i)\}_{i=1}^{n}$, which can be an indicator of the possibility of a cluster to be an outlier cluster \cite{liu2025flexuod}. Specifically, we adopt the Spearman’s rank correlation coefficient \cite{fieller1961tests} to compute the ranking-index similarity \cite{liu2025flexuod} between $\{f_{t}(x_i)\}_{i=1}^{n}$ and $\{f_{base}(x_i)\}_{i=1}^{n}$:
\begin{equation}
\begin{aligned}
\phi_{t, base} = |\texttt{Spearman}(\operatorname{rank-index}(\{f_{t}(x_i)\}_{i=1}^{n}),
\operatorname{rank-index}(\{f_{base}(x_i)\}_{i=1}^{n})|,
\label{equ 5}
\end{aligned}
\end{equation}
where $|\cdot|$ represents the absolute value function.
To align the value of $\phi_{t, base}$ across different domains, we apply the $\operatorname{min-max}$ normalization to it.
If $\phi_{t, base}$ is less than a small positive value $\epsilon^{+}$, we will remove $\mathbf{C}_t$ from $\mathbf{X}$, and the refined outlier score function is defined as:
\begin{equation}
f_{refine}(x_i) = \displaystyle || x_i - m_{\mathbf{X}_{in}^{\prime}} ||^{2}, \mathbf{X}_{in}^{\prime} = \{ \mathbf{X} \setminus \mathbf{C}_t | \phi_{t, base}< \epsilon^{+} \},
\label{equ 6}
\end{equation}
and $\epsilon^{+}$ is fixed to $0.9$ to obtain uncontaminated inlier features. As visualized in Fig. \ref{fig 6}, the refined distance transform
(Eq. \ref{equ 6}) significantly outperforms the baseline (Eq. \ref{equ 4}) across varying outlier ratios.

\subsection{Implementation}
\label{sec 4.3}
\noindent \textbf{Pseudo Labeling.}
Inspired by the above analysis, we can further compare each $f_t(x_i) = \displaystyle || x_i - {m_{\mathbf{C}_{t}}} ||^{2}$ with $f_{refine}$:
\begin{equation}
\begin{aligned}
\phi_{t,refine} = |\texttt{Spearman}(\operatorname{rank-index}(\{f_{t}(x_i)\}_{i=1}^{n}),
\operatorname{rank-index}(\{f_{refine}(x_i)\}_{i=1}^{n})|.
\label{equ 7}
\end{aligned}
\end{equation}
If $\phi_{t, refine}$ is large, we consider the $t$-th cluster as an inlier cluster; if $\phi_{t, refine}$ is relatively small, we consider it an outlier cluster.
Based on the basic definition of the UOD task, outliers may derive from various classes of data distribution that are out of the normal range. Besides, although inliers usually come from one single class, we argue that it also demonstrates high within-class variation. For example, the “airplane” class not only involves standard airplanes but also has some wings, passengers, or cockpits. Such that in this work, we define two decision boundaries, $\epsilon^{+}$ is set to $0.9$ while $\epsilon^{-}$ is set to $0.5$, respectively, to maintain the balance between sample quantity and purity of both inliers and outliers. The predicted inlier and outlier supervision in the second \textcolor{red}{classifying} phase is denoted as $\mathbf{X}_{in}^{\prime} = \{\mathbf{C}_t | \phi_{t,refine} > \epsilon^{+} \}$ and $\mathbf{X}_{out}^{\prime} = \{\mathbf{C}_t | \phi_{t, refine} < \epsilon^{-} \}$.

\begin{figure}[htbp]
\centering
\includegraphics[width=0.5\textwidth]{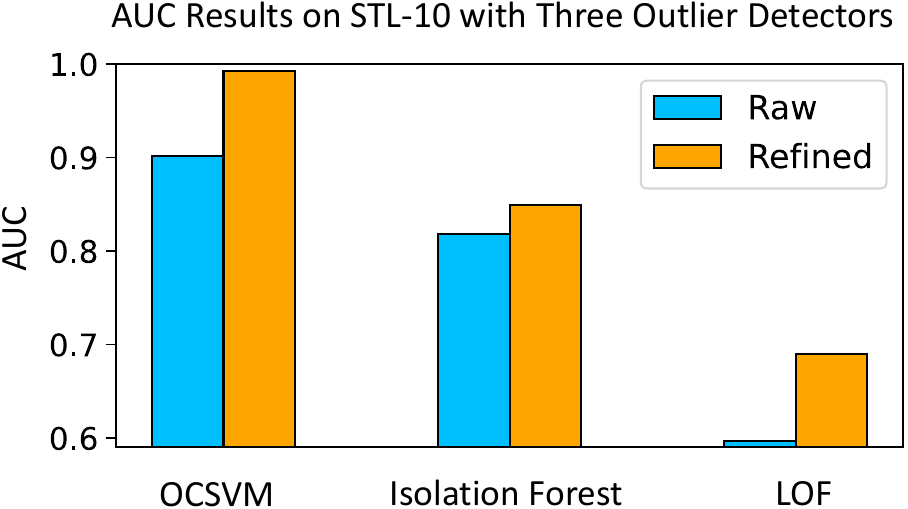}
\caption{\textbf{The AUC results comparison between the refined feature and the raw feature with three outliers.}
The refined feature can significantly improve the detection accuracy of three outlier detectors.
}
\label{fig 7} 
\end{figure}

\begin{figure*}[ht]
\centering
\includegraphics[width=0.95\textwidth]{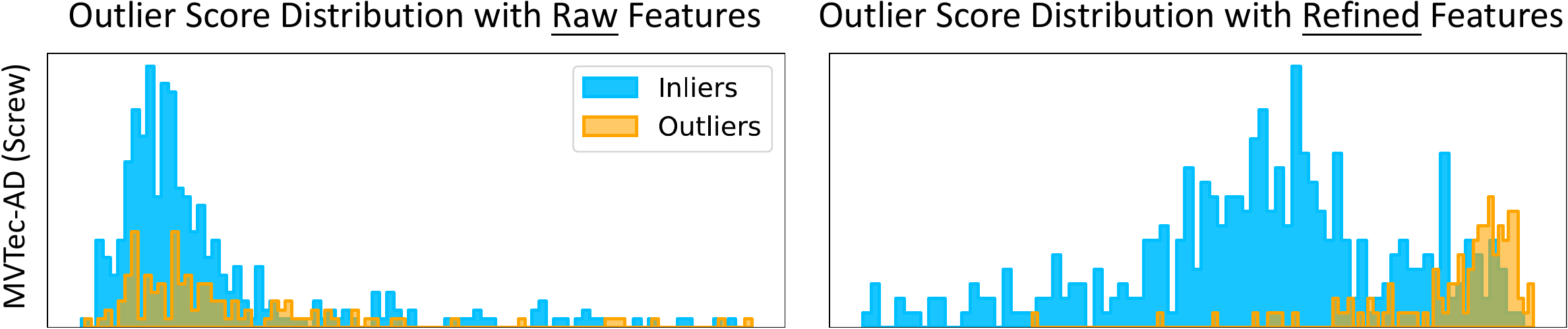}
\caption{\textbf{The histogram visualization of outlier score distribution with both raw features and enhanced features.} The enhanced features exhibit substantially improved separability between inliers and outliers compared with the raw features.}
\label{fig 8} 
\end{figure*}

\noindent \textbf{Contrastive Learning.}
Then, we are able to leverage the contrastive loss to enlarge the gap between inlier and outlier features as well as reduce the variation within inlier features. During each training epoch, we randomly sample positive pairs $(x_i, x_j)$ from $\mathbf{X}_{in}^{\prime}$ and negative pairs $(x_i, x_j)$ from the combination of $\mathbf{X}_{in}^{\prime}$ and $\mathbf{X}_{out}^{\prime}$, corresponding to similar pairs with label $y=0$ and dissimilar pairs with label $y=1$, respectively. To avoid memory explosion and improve training stability, pairs are dynamically resampled at each epoch rather than generating all possible pairs at once.
We define a simple two-layer perceptron (MLP) contrastive learning model $g_\theta: \mathbb{R}^d \to \mathbb{R}^{d^{\prime}}$ consisting of two linear layers with ReLU activation, followed by $l_2$ normalization of the output embedding:
\begin{equation}
\begin{aligned}
g_\theta(x) = \texttt{normalize}\Big(\mathbf{W}_2 , \mathrm{ReLU}(\mathbf{W}_1 x + b_1) + b_2 \Big).
\end{aligned}
\label{equ 8}
\end{equation}
Given a sample pair $(x_i, x_j)$ and its predicted label $y \in \{0,1\}$, we adopt the following contrastive loss:
\begin{equation} 
\begin{aligned}
\mathcal{L}(x_i, x_j, y) &  = (1-y) \cdot || g_\theta(x_i) - g_\theta(x_j) ||_2^2 \\
& + y \cdot  \max\big(0, m - || g_\theta(x_i) - g_\theta(x_j) ||_2\big)^2,
\end{aligned}
\label{equ 9}
\end{equation}
where $m$ is a margin ($1.0$) hyperparameter that enforces a minimum distance between negative pairs. During training, the model performs forward propagation on the dynamically sampled positive and negative pairs to compute the loss $\mathcal{L}$, and the parameters $\theta \leftarrow \{\mathbf{W}_1, b_1, \mathbf{W}_2, b_2 \}$ are updated via the Adam optimizer \cite{kingma2014adam}.
In inference, all feature samples $\mathbf{X}$ are passed through the trained model to obtain embeddings $\mathbf{Z} = g_{\theta}(\mathbf{X})$, which are then normalized.
Consequently, we feed $\mathbf{Z}$ into the classic OCSVM \cite{scholkopf2001estimating} to predict the outlier scores $f(x_i) = \operatorname{OCSVM}(z_i \in \mathbf{Z}) \leftarrow g_{\theta}(x_i \in \mathbf{X})$. We select OCSVM as the outlier scoring head because OCSVM leverages the kernel trick to define complex, non-linear decision boundaries in high-dimensional feature spaces, making it more sensitive to subtle anomalousness. As validated in Fig. \ref{fig 7}, our refined feature can significantly improve the detection accuracy (AUC score) of three classic outlier detectors, while OCSVM shows the best performance. As demonstrated in Fig. \ref{fig 8}, for a hard case, the outlier score discriminativeness of our VUOD framework will be more separable than VUOD with the raw feature representation. 

\section{Evaluation}
\label{sec:evaluation}

\subsection{Experimental Settings}
\noindent \textbf{Benchmark Datasets.}
The evaluated benchmark datasets involve $\operatorname{(\romannumeral1)}$ Natural images: STL-10 \cite{coates2011analysis}, Internet \cite{lin2021shell}, Caltech-101 \cite{fei2006one}, CIFAR-10 \cite{krizhevsky2009learning}, CIFAR-100 \cite{krizhevsky2009learning}, MIT-Places \cite{zhou2017places}; $\operatorname{(\romannumeral2)}$ Industrial images: MVTec-AD \cite{bergmann2019mvtec}, BTAD \cite{mishra2021vt}, MPDD \cite{jezek2021deep} and MVTecLOCO \cite{bergmann2022beyond}; $\operatorname{(\romannumeral3)}$ Medical images: LiverCT, BrainMRI, OCT and RESC, from a medical imaging benchmark BMAD \cite{bao2024bmad}.

\noindent \textbf{Building A Target Dataset.}
For natural image datasets, we adopt ResNet-50 \cite{he2016deep} and CLIP \cite{radford2021learning} as two default feature extractors, and the target dataset is made up of all inliers from one class and some randomly selected outliers from other classes. There are six outlier ratios ranging from $0.01$ to $0.5$ \cite{liu2025flexuod}.
For other benchmarks, we combine all samples from both training and test set and utilize ResNet-18 \cite{he2016deep} and Wide ResNet-101 \cite{zagoruyko2016wide}, followed by the existing work \cite{batzner2024efficientad}.

\noindent \textbf{Competing Baselines.}
We mainly compare with a series of existing UOD methods: Deep SVDD \cite{ruff2018deep}, RSRAE \cite{lai2020rsrae}, GOAD \cite{bergman2020goad}, Shell-Re \cite{lin2021shell}, NeuTraL \cite{qiu2021neural}, ICL \cite{shenkar2021anomaly}, LUNAR \cite{lai2020rsrae}, ECOD \cite{li2023ecod}, LVAD \cite{lin2022locally}, SLAD \cite{xu2023fascinating}, DeepIF \cite{xu2023deep}, Multi-T \cite{liu2024rethinking} and FlexUOD \cite{liu2025flexuod}. And for the industrial MVTec-AD evaluation, we also compare with some specific industrial defect detectors: PaDiM \cite{defard2021padim}, FRE \cite{ndiour2023fre}, and EfficientAD \cite{batzner2024efficientad}. These approaches are also trained and tested on the same target dataset (involving both normal data and defective data).    

\noindent \textbf{Evaluation Metric.}
We mainly evaluate the ranking accuracy of the predicted outlier scores using the AUROC (AUC) score. 

\begin{table*}[htbp]
\scriptsize
\renewcommand{\tabcolsep}{1.5pt}
\renewcommand{\arraystretch}{1.1}
\centering
\caption{\text{AUC results on a series of natural image datasets.} \textcolor{red}{\textbf{Red}} and \textcolor{blue}{\textbf{Blue}} indicates the best and second-best results, respectively.} 
	{
\begin{tabular}{l|c|c|c|c|c|c|c}
\hline
 \multirow{1}{*}{\scriptsize{Method}} & \multirow{1}{*}{\scriptsize{Venue}} & \multicolumn{1}{c|}{STL-10} & \multicolumn{1}{c|}{Internet}   & \multicolumn{1}{c|}{Caltech-101}   & \multicolumn{1}{c|}{CIFAR-10} & \multicolumn{1}{c|}{CIFAR-100} & \multicolumn{1}{c}{MIT-Places} \\
\hline
Deep SVDD \cite{ruff2018deep} & ICML-2018  & 0.593  & 0.674 & 0.745 & 0.533 & 0.575 & 0.539 \\
RSRAE \cite{lai2020rsrae} & ICLR-2019  & 0.903  & 0.916 & \textcolor{blue}{\textbf{0.986}}  & 0.816 & 0.889 & 0.778   \\
GOAD \cite{bergman2020goad} & ICLR-2020 & 0.946 & {{0.945}} & 0.981 & 0.862 & 0.886 & {{0.871}}  \\
Shell-Re \cite{lin2021shell} & TPAMI-2021  & 0.866  & 0.896 & 0.905  & 0.867   & 0.834 & 0.822   \\
NeuTraL \cite{qiu2021neural}  & ICML-2021 & 0.828  & 0.877 & 0.633  & 0.748  & 0.815  & 0.711 \\
ICL \cite{shenkar2021anomaly} & ICLR-2022 & 0.929  &  {0.937} & 0.964  & 0.859  & {{0.911}} & 0.832 \\
LUNAR \cite{goodge2022lunar}  & AAAI-2022 & 0.781  & 0.776 & 0.923  & 0.766  & 0.859  & 0.698 \\
ECOD \cite{li2023ecod} & TKDE-2022  & 0.932  & 0.925 & 0.976  & {{0.888}}  & 0.905 & 0.846  \\
LVAD  \cite{lin2022locally} & ECCV-2022 & {{0.948}} & 0.923 & 0.977  & 0.864  & 0.907  &  {0.860} \\
SLAD  \cite{xu2023fascinating} & ICML-2023 & 0.923  & 0.919  & 0.888  & 0.861  & 0.899  & 0.846  \\
DeepIF \cite{xu2023deep} & TKDE-2023  & 0.858  & 0.866  & 0.874  & 0.801  & 0.808  & 0.765  \\
Multi-T \cite{liu2024rethinking} &  ECCV-2024   & {{0.957}}  & {{0.956}}  & 0.985  & {{0.888}}  & {{0.917}} & 0.859    \\
FlexUOD \cite{liu2025flexuod} &  CVPR-2025 & \textcolor{blue}{\textbf{0.980}}  & \textcolor{blue}{\textbf{0.981}} & {{0.983}} & \textcolor{blue}{\textbf{0.942}} & \textcolor{blue}{\textbf{0.947}} & \textcolor{blue}{\textbf{0.928}}   \\
\hline
VUOD (Ours) &  $-$  
& \textcolor{red}{\textbf{0.994}}  & \textcolor{red}{\textbf{0.991}}  &  \textcolor{red}{\textbf{0.994}} &  \textcolor{red}{\textbf{0.972}} & \textcolor{red}{\textbf{0.981}}  & \textcolor{red}{\textbf{0.971}}   \\
\hline
\end{tabular}}
\label{table 1} 
\end{table*}

\begin{table*}[t]
\scriptsize
\renewcommand{\tabcolsep}{0.2pt}
\renewcommand{\arraystretch}{1.1}
\centering
\caption{\text{AUC results on industrial inspection and medical imaging (disease detection).} The best
results are highlighted in bold.}
 \begin{tabular}{c|l|ccccc|ccccc}
\hline
\multirow{2}{*}{Type} & \multirow{2}{*}{Dataset} & \multicolumn{5}{c|}{ResNet-18} & \multicolumn{5}{c}{Wide ResNet-101}  \\
\cline{3-12} 
& & RSRAE  & LVAD   &  Multi-T   &  FlexUOD  &  VUOD   & RSRAE  & LVAD   &  Multi-T   &  FlexUOD  &  VUOD\\
\hline 
\multirow{4}*{\rotatebox{90}{Industrial}}
& MVTec-AD    & 0.709 & 0.779 & 0.757 & 0.752 & \textbf{0.911} & 0.770 & 0.799 & 0.780 & 0.778 & \textbf{0.931} \\
& BTAD   & 0.873 & 0.894 & 0.885 & 0.882 & \textbf{0.902} & 0.846 & 0.884 & 0.886 & 0.883 & \textbf{0.928} \\
& MPDD    & 0.618 & 0.573 & 0.563 & 0.564 & \textbf{0.906} & 0.592 & 0.602 & 0.603 & 0.595 & \textbf{0.926}  \\
& MVTec-LOCO   & 0.655 & 0.693 & 0.661 & 0.663 & \textbf{0.812} & 0.629 & 0.695 & 0.657 & 0.660 & \textbf{0.832}  \\
\hline
\multirow{4}*{\rotatebox{90}{Medical}}
& LiverCT      & 0.666 & 0.663 & 0.696 & 0.708 & \textbf{0.813} & 0.703 & 0.571 & 0.652 & 0.567 & \textbf{0.838}  \\
& BrainMRI    & 0.541 & 0.623 & 0.583 & 0.584 & \textbf{0.804} & 0.676 & 0.647 & 0.652 & 0.578 & \textbf{0.790}   \\
& OCT     & 0.524 & 0.778 & 0.730 & 0.612 & \textbf{0.880} & 0.713 & 0.811 & 0.725 & 0.588 & \textbf{0.877}  \\
& RESC    & 0.712 & 0.810 & 0.823 & \textbf{0.948} & 0.901 & 0.662 & 0.746 & 0.644 & 0.589 & \textbf{0.931}  \\
\hline
\end{tabular}
\label{table 2} 
\end{table*}

\begin{table}[ht]
\scriptsize
\renewcommand{\tabcolsep}{1.8pt}
\renewcommand{\arraystretch}{1.1}
\centering
\caption{\text{AUC results for each category on MVTec-AD.}}
 \begin{tabular}{c|l|ccc||cc}
\hline
\multirow{2}{*}{Type} & Category & PaDiM \cite{defard2021padim}  & FRE \cite{ndiour2023fre}  & EfficientAD \cite{batzner2024efficientad}  & VUOD & VUOD  \\
\cline{2-7}
 & Backbone & ResNet-18 & ResNet-50 & Wide ResNet-101 & ResNet-18 & Wide ResNet-101 \\
\hline
\multirow{5}*{\rotatebox{90}{Textures}}
 & Carpet & 0.986 & 0.968 & 0.969 & 0.875 & 0.901\\
 & Grid & 0.865 & 0.869 & 0.996 & 0.816 & 0.859\\
 & Leather & 0.997 & 1.000 & 0.976 & 0.841 & 0.971 \\
 & Tile & 0.939 & 0.994 & 0.960 & 0.924 & 0.972 \\
 & Wood & 0.987 & 0.992 & 0.903 & 0.939 & 0.991 \\
\hline
\multirow{10}*{\rotatebox{90}{Objects}}
 & Bottle & 0.996 & 0.993 & 1.000 & 0.927 & 0.882  \\
 & Cable & 0.789 & 0.892 & 0.852 & 0.929 & 0.908 \\
 & Capsule & 0.851 & 0.744 & 0.753 & 0.931 & 0.948\\
 & Hazelnut & 0.852 & 0.940 & 0.766 & 0.949 & 0.946  \\
 & Metal Nut & 0.928 & 0.914 & 0.977 & 0.949 & 0.948  \\
 & Pill & 0.764 & 0.838 & 0.892 & 0.891 & 0.885  \\
 & Screw & 0.811 & 0.789 & 0.940 & 0.903 & 0.921 \\
 & Toothbrush & 0.958 & 0.931 & 0.970 & 0.956 & 0.977\\
 & Transistor & 0.908 & 0.894 & 0.853 & 0.929 & 0.908 \\
 & Zipper & 0.896 & 0.942 & 0.956 & 0.909 & 0.944\\
\hline
 & Average & 0.902 & 0.913 & 0.917 &  {0.911} &  {\textbf{0.931}} \\
\hline
\end{tabular}
\label{table 3} 
\end{table}

\subsection{Result Analysis}
Table \ref{table 1} shows the significant improvements compared with several competitive baselines across a diverse range of outlier ratios, two feature extractors, and various benchmark datasets. Notably, even with the challenging datasets such as CIFAR-10 and CIFAR-100 \cite{perera2019ocgan}, our method outperforms current SOTA AUC scores by margins of about $3$\% and $4$\%, respectively. 
Besides, we observe a $5\%$ performance improvement on the scene understanding/recognition dataset, MIT-Places, increasing the AUC score from $0.928$ to $0.971$, which indicates the potential application of VUOD in enhancing performance on related tasks. 

In Table \ref{table 2}, we demonstrate the results on benchmark datasets that go beyond natural images, such as industrial defect detection datasets (e.g., MVTec-AD, MVTec-LOCO) and medical disease detection datasets (e.g., LiverCT and BrainMRI) that are usually patch/pixel-level contaminated. To maintain the evaluation consistency with Table \ref{table 1}, we still use the default feature embedding of the entire image, which does not seem like the optimal choice, but VUOD still significantly outperforms the baselines. Without any supervision, its AUC score exceeds $0.910$ on industrial datasets. 
In Table~\ref{table 3}, we compare VUOD with several recent industrial defect detection methods on MVTec-AD. Remarkably, although VUOD is not specifically designed for this task, it still delivers competitive results. Moreover, it supports ResNet-18 as a reliable backbone, which offers a significant advantage for practical deployment.

\begin{table}[htbp] 
\scriptsize
\renewcommand{\tabcolsep}{0.6pt}
\renewcommand{\arraystretch}{1.1}
\centering
\caption{Timing (Seconds) is measured with 10,000 ResNet-50 feature samples. \#$\operatorname{AP}$ refers to the automatic defined clustering number of $\operatorname{AP}$.}
\begin{tabular}{l|c|c|c}
\hline
$\operatorname{AP}$ & $k$-Means ($k=10$) & $k$-Means ($k=100$)  & $k$-Means ($k=$\#$\operatorname{AP}$) \\
\hline
 \cellcolor{gray!20}{84.60} & 9.588 &  61.20 & 207.3 \\ 
\hline
\end{tabular}
\label{table 4} 
\end{table}

\begin{table}[htbp] 
\scriptsize
\renewcommand{\tabcolsep}{0.6pt}
\renewcommand{\arraystretch}{1.1}
\centering
\caption{Efficiency between VUOD and EfficientAD on MVTec-AD. We evaluate efficiency on the ``bottle'' category (input size: 256$\times$256). Both methods use Wide ResNet-101 as the backbone and are conducted on one NVIDIA RTX 3090 GPU.}
\begin{tabular}{l|ccc}
\hline
Method & Timing (Seconds) & FLOPs (G) & GPU Memory (MB) \\
\hline
EfficientAD & 42.10 & 75.91 & 102.0 \\
\cellcolor{gray!20}{VUOD} & \cellcolor{gray!20}{\textbf{17.18}} & \cellcolor{gray!20}{\textbf{22.80}} & \cellcolor{gray!20}{\textbf{39.23}} \\
\hline
\end{tabular} 
\label{table 5} 
\end{table}

\noindent \textbf{Efficiency Analysis.} We first conduct the efficiency comparison between $\operatorname{AP}$ and $k$-Means in Table~\ref{table 4} since it is the most important component of VUOD. Notably, the running time of $k$-Means depends on the cluster number $k$, which is known to be challenging to determine without any supervision. When $k$ is set to the same as $\operatorname{AP}$, $\operatorname{AP}$ runs more quickly. This occurs because our clustering tasks involve small sample sizes, where $\operatorname{AP}$’s quadratic message updates remain inexpensive. In contrast, $k$-Means performs multiple random initializations, each requiring high-dimensional distance computations, and often needs to be executed with several candidate values of $k$. $\operatorname{AP}$ avoids these costs by running a single deterministic optimization and automatically determining the number of clusters. Consequently, $\operatorname{AP}$ becomes empirically faster than $k$-Means in small-scale and high-dimensional feature clustering, as evaluated in Table \ref{table 4}. 
In Table \ref{table 5}, our solution not only excels in detection accuracy but also efficiency on the industrial inspection dataset MVTec-AD, compared with EfficientAD \cite{batzner2024efficientad}.

\begin{table}[htbp] 
\scriptsize
\renewcommand{\tabcolsep}{0.6pt}
\renewcommand{\arraystretch}{1.1}
\centering
\caption{\text{Ablation study about VUOD.} The results are AUC scores of OCSVM with the raw and enhanced features.}
 \begin{tabular}{l|ccccccc} 
\hline
Method & STL-10 & Internet & Caltech-101 & CIFAR-10  & CIFAR-100  & MIT-Places & MVTec-AD \\
\hline
OCSVM (Raw Feat.)      &  0.899  &  0.903   &  0.955   &   0.824   &  0.845 & 0.824  & 0.754 \\
OCSVM (Re. Feat.)      &  0.994  &  0.991   &  0.994   &   0.972   &  0.981  & 0.971  & 0.921\\
\cellcolor{gray!20}{Improvement}            &  \cellcolor{gray!20}{10.6\%} &   \cellcolor{gray!20}{9.75\%}   &  \cellcolor{gray!20}{4.08\%}   & \cellcolor{gray!20}{18.0\%}    &  \cellcolor{gray!20}{16.1\%}    &  \cellcolor{gray!20}{17.8\%}   &    \cellcolor{gray!20}{22.1\%} \\ 
\hline
\hline
Method & BTAD  & MPDD   & MVTec-LOCO  & LiverCT & BrainMRI & OCT & RESC \\
\hline
OCSVM (Raw Feat.)   &  0.874   &  0.575   &  0.662   &  0.597   &   0.605  &  0.747   &  0.701  \\
OCSVM (Re. Feat.)   &  0.915   &  0.916   &    0.822   &  0.825   &  0.797  &  0.879  &  0.916 \\
\cellcolor{gray!20}{Improvement}              &  \cellcolor{gray!20}{4.69\%}   &   \cellcolor{gray!20}{59.3\%}  &  \cellcolor{gray!20}{24.2\%}   &  \cellcolor{gray!20}{38.2\%}   &  \cellcolor{gray!20}{31.7\%}   &  \cellcolor{gray!20}{17.7\%}    &  \cellcolor{gray!20}{30.7\%} \\ 
\hline
\end{tabular}
\label{table 6} 
\end{table}

\begin{table}[htbp]
\scriptsize
\renewcommand{\tabcolsep}{3.5pt}
\renewcommand{\arraystretch}{1.1}
\centering
\caption{Ablation study about parameter setting ($\epsilon^{+}$, $\epsilon^{-}$) in Sec. \ref{sec 4.3}. The results are AUC scores of VUOD on CIFAR-10.} 
\begin{tabular}{l|c|c|c|c|c|c|c|c} 
\hline
Method & (0.9, 0.7)  & (0.9, 0.6) &  {(0.9, 0.5)} & (0.9, 0.4)  & (0.8, 0.6) & (0.8, 0.5)  & (0.8, 0.4)  & (0.8, 0.3) \\
\hline
VUOD &  0.969  &  0.971 &  \cellcolor{gray!20}{\textbf{0.972}} &  0.970 &  0.968   &  0.969  &   0.968  &  0.965   \\
\hline
\hline
Method & (0.7, 0.5)  & (0.7, 0.4) & (0.7, 0.3) & (0.7, 0.2)  & (0.6, 0.4) & (0.6, 0.3)  & (0.6, 0.2)  & (0.6, 0.1) \\
\hline
VUOD & 0.970    &  0.968  &  0.967  &  0.964 &   0.967  & 0.966   & 0.964  & 0.963   \\
\hline
\end{tabular}
\label{table 7}  
\end{table}

\subsection{Ablation Study}
To measure the feature discriminativeness enhancement contribution of the VUOD framework, we conduct ablation studies on all benchmark datasets, and AUC results are averaged on two feature extractors.
Compared with the default OCSVM, our feature enhancement techniques improve the performance up to $59.3\%$, as shown in Table \ref{table 6}. 
For all evaluated benchmarks, we used a consistent set of default parameters ($\epsilon^{+}$, $\epsilon^{-}$) in Sec. \ref{sec 4.3} without per-category tuning. To further validate this, we conducted an additional robustness study (Table \ref{table 7}), shifting the default thresholds by $0.2$ resulted in only a marginal performance fluctuation ($\leq 0.5\%$ in AUC score), suggesting that they are not sensitive to specific values. 

\begin{table}[htbp]
\scriptsize
\renewcommand{\tabcolsep}{2.5pt}
\renewcommand{\arraystretch}{1.1}
\centering
\caption{\text{Image classification results on Imagenette \cite{imagenette2019}, a smaller set of ImageNet \cite{deng2009imagenet}.} The image classifier baseline we employed is ResNet-18.} 
\begin{tabular}{l|c|c} 
\hline
Method & Top-1 Accuracy  & Top-5 Accuracy  \\
\hline
Baseline &  90.81\%   &  99.11\%  \\
\hline 
\cellcolor{gray!20}{w/ VUOD}  &   \cellcolor{gray!20}{\textbf{91.45\%}}    &  \cellcolor{gray!20}{\textbf{99.39\%}} \\
\hline
\end{tabular}
\label{table 8}  
\end{table}

\begin{figure*}[ht]
\centering
\includegraphics[width=0.95\textwidth]{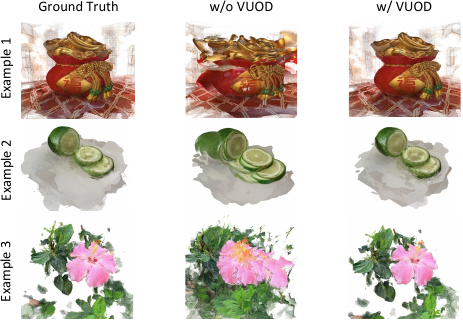} 
\caption{\textbf{3D Dense Point Map Reconstruction Results of VGGT \cite{wang2025vggt}.}
Ground Truth refers to the reconstruction obtained using VGGT on an uncontaminated image set. Three target datasets are visualized in the Appendix.
}
\label{fig 9} 
\end{figure*} 

\subsection{Diverse Visual Applications} 
\label{sec. 5.4}
The above standard UOD evaluation focuses on detecting outliers from various domains, but lacks applying this mechanism to other topics. To further validate the versatile ability of our VUOD framework, we consider it as an image-level data pre-processor to filter anomalous samples from the training/input set of two important visual tasks: image classification and 3D reconstruction.   
And we compute the threshold $\tau$ based on the predicted outlier ratio $\gamma$ from FlexUOD\cite{liu2025flexuod}, i.e., $ \tau = \operatorname{sort}(\{f(x_i)\}_{i=1}^{n})[n\cdot(1-\gamma)]$. If the predicted outlier score $f(x_i)> \tau$, $x_i$ will be classified as an outlier. Otherwise, it belongs to the inliers.

\noindent \textbf{Image Classification.}
By integrating VUOD into the classification pipeline, VUOD can effectively identify and remove outliers from the default training set, which naturally involves some noisy images \cite{yun2021re}, thereby improving the classification accuracy of ResNet-18, as shown in Table \ref{table 8}. 

\noindent \textbf{3D Reconstruction.}
We further extend VUOD to 3D reconstruction, specifically within the VGGT \cite{wang2025vggt} framework. As VGGT requires a sequence of images, it is inherently sensitive to image contamination. The absence of an outlier filtering mechanism leads to significant reconstruction artifacts, illustrated in Fig. \ref{fig 9}. We compare our solution against the latest RobustVGGT \cite{han2026emergent}, which also employs image-level rejection but is tailored for VGGT. Notably, RobustVGGT requires a time-consuming preliminary pass through the VGGT pipeline to extract internal representations. In contrast, VUOD is feature-agnostic, supporting flexible backbones such as ResNet and CLIP. In experiments across three target datasets in Fig. \ref{fig 9}, both two approaches successfully identified all outliers. However, RobustVGGT requires $118$ ms per image, VUOD reduces processing time to just $27$ ms, achieving a $4.4\times$ speedup on an NVIDIA RTX 4090D GPU. 

\subsection{Limitation}
Since VUOD aims to be available for various UOD domains, it operates at the whole-image level and is designed to rank samples based on their outlier scores, it may be less effective for patch-level or pixel-level outlier segmentation tasks \cite{chan2021segmentmeifyoucan}.

\section{Conclusion}
\label{sec:conclusion}
In this work, we aim to break the application limitation of traditional unsupervised outlier detection by introducing the VUOD framework. It is designed to improve feature discriminativeness via pseudo labeling and contrastive learning. Our evaluation demonstrates that VUOD offers promising performance and stability across not only standard evaluation but also diverse applications such as industrial inspection and medical imaging.
Additionally, VUOD is distinguished by its plug-and-play advantage that facilitates easy integration with other critical computer vision tasks like image classification and 3D reconstruction, to improve their classification accuracy and reconstruction robustness.  

\section*{Acknowledgement}
This work is partially supported by Shenzhen Science and Technology Program KQTD20210811090149095.


%
%
\bibliographystyle{splncs04}
\bibliography{main}
\end{document}